\documentclass[runningheads]{llncs}

\usepackage{eccv}

\usepackage{eccvabbrv}
\usepackage{graphicx}
\usepackage{booktabs}

\usepackage[accsupp]{axessibility}  

\usepackage{hyperref}

\usepackage{orcidlink}

\usepackage{amsmath}
\usepackage{graphicx}
\usepackage{mathtools}
\usepackage{enumitem}
\usepackage{colortbl}
\usepackage[capitalize]{cleveref}
\usepackage{amssymb}
\usepackage{pifont}
\usepackage{multirow}
\usepackage{wrapfig}
\usepackage{etoolbox}
\BeforeBeginEnvironment{wrapfigure}{\setlength{\intextsep}{0pt}}

\newcommand{\name}{\texttt{CoMusion}}

\definecolor{gray}{rgb}{0.85,0.85,0.85}
\newcommand{\cmark}{\ding{51}}
\newcommand{\xmark}{\ding{55}}

\usepackage{amsmath,amsfonts,bm}

\def\eqref#1{equation~\ref{#1}}

\def\1{\bm{1}}

\DeclareMathAlphabet{\mathsfit}{\encodingdefault}{\sfdefault}{m}{sl}
\SetMathAlphabet{\mathsfit}{bold}{\encodingdefault}{\sfdefault}{bx}{n}

\def\gL{{\mathcal{L}}}

\def\gN{{\mathcal{N}}}

\def\sR{{\mathbb{R}}}

\newcommand{\E}{\mathbb{E}}

\begin{document}

\title{\texttt{CoMusion}: Towards Consistent Stochastic Human Motion Prediction via Motion Diffusion} 

\titlerunning{\name}

\author{Jiarui Sun\orcidlink{0000-0002-7113-081X} \and
Girish Chowdhary\orcidlink{0000-0002-4657-307X}}

\authorrunning{J.~Sun and G.~Chowdhary}

\institute{University of Illinois Urbana-Champaign \\
\email{\{jsun57,girishc\}@illinois.edu}}

\maketitle

\begin{abstract}

Stochastic Human Motion Prediction (HMP) aims to predict multiple possible future human pose sequences from observed ones. 
Most prior works learn motion distributions through encoding-decoding in the latent space, which does not preserve motion’s spatial-temporal structure.
While effective, these methods often require complex, multi-stage training and yield predictions that are inconsistent with the provided history.
To address these issues, we propose \name, a single-stage, end-to-end diffusion-based stochastic HMP framework.
\name~is inspired from the insight that a smooth future pose initialization improves prediction performance, a strategy not previously utilized in stochastic models but evidenced in deterministic works.
To generate such initialization, \name's motion predictor starts with a Transformer-based network for initial reconstruction of corrupted motion. 
Then, a graph convolutional network (GCN) is employed to refine the prediction considering past observations in the discrete cosine transformation (DCT) space. 
Our method, facilitated by the Transformer-GCN module design and a proposed variance scheduler, excels in predicting accurate, realistic, and consistent motions, while maintaining appropriate diversity.
Experimental results on benchmark datasets demonstrate that \name~surpasses prior methods across metrics, while demonstrating superior generation quality. 
Code is released at \url{https://github.com/jsun57/CoMusion/}.
\end{abstract}

\section{Introduction}
\label{sec::intro}

Human Motion Prediction (HMP) aims to forecast human movements based on observed motion trajectories.
This task has a wide range of applications \cite{DBLP:conf/cvpr/ZhangBT21, yang2023neural, DBLP:conf/iclr/0007K20, xu2023interdiff, xu2023stochastic, DBLP:conf/cvpr/BhattacharyyaSF18, liu2022intention, troje2002decomposing, ju2023humanart}, spanning autonomous driving \cite{DBLP:journals/tiv/PadenCYYF16}, robotics \cite{DBLP:conf/iros/GuiZWLMV18}, animation creation \cite{DBLP:journals/cgf/WelbergenBERO10}, and healthcare \cite{DBLP:journals/sensors/TaylorSDZAI20}.
A considerable body of research tackles the deterministic HMP problem, aiming to predict a single, most probable future pose sequence \cite{DBLP:conf/eccv/MaoLS20, DBLP:conf/cvpr/MaNLZL22, DBLP:conf/iccv/DangNLZL21}.
Among these works, the top-performing models \cite{DBLP:conf/cvpr/MaNLZL22, li2022skeleton} demonstrate that graph convolutional networks (GCN) is very suitable
for HMP.
By coupling GCN with discrete cosine transformation (DCT), these GCN-DCT methods treat human poses as graphs and explicitly model spatial-temporal relations among joints, which benefits motion prediction.
However, the deterministic methods fall short in contexts such as autonomous driving in crowded areas, where predicting different possible human motions is crucial.

\begin{wrapfigure}[16]{R}{0.5\textwidth}
\centering
\includegraphics[width=0.5\textwidth]{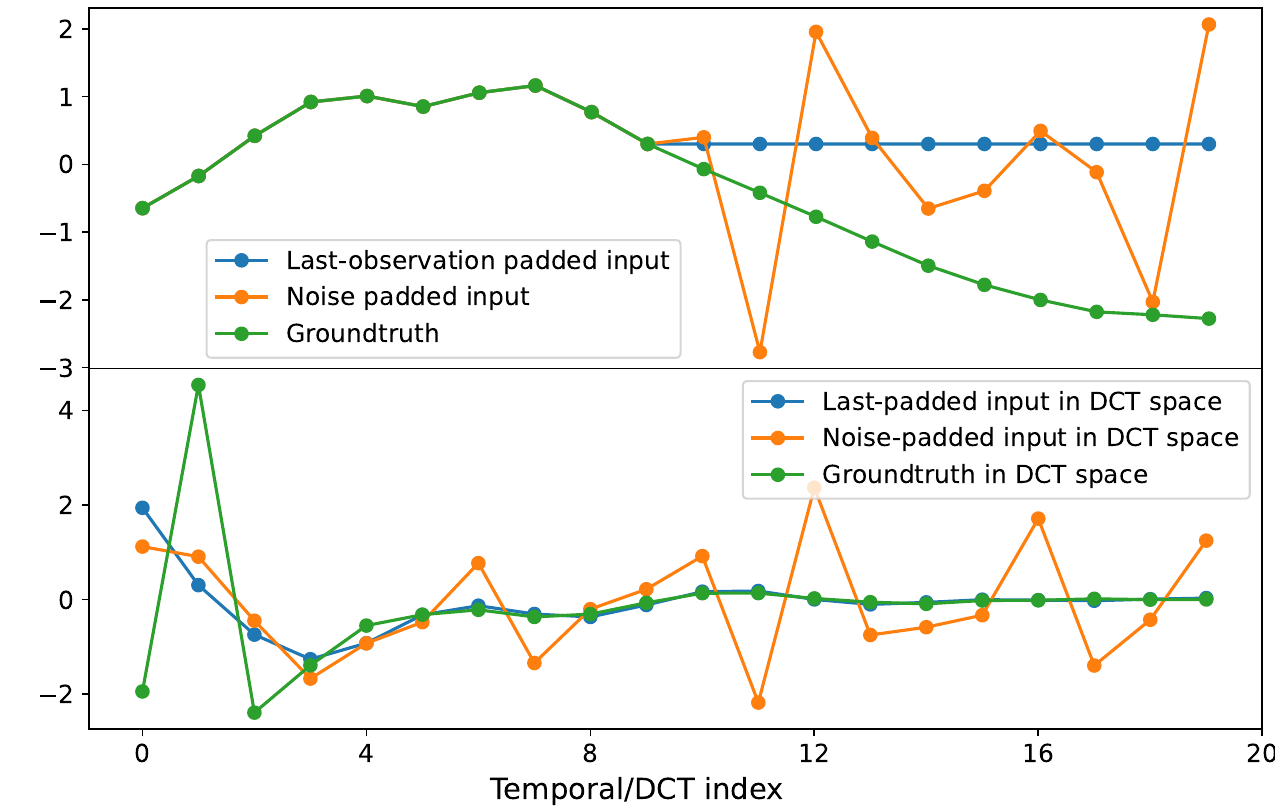}
\caption{\textbf{Top}: Three joint motion trajectories (length 20), last 10 features vary among the last-observation-padded, noise-padded and groundtruth sequences. \textbf{Bottom}: Their corresponding DCT values.}
\label{fig:dct}
\end{wrapfigure}

Recent research has shifted toward the stochastic paradigm of HMP, adopting generative models to learn conditional motion distributions \cite{DBLP:conf/nips/GoodfellowPMXWOCB14, DBLP:conf/cvpr/BarsoumKL18, DBLP:conf/aaai/KunduGB19, DBLP:journals/corr/KingmaW13, DBLP:conf/eccv/YuanK20, DBLP:conf/iccv/MaoLS21, DBLP:conf/icml/Sohl-DicksteinW15, barquero2023belfusion, DBLP:conf/aaai/WeiSLLLSH23, chen2023humanmac}. 
These stochastic approaches, while effective in certain scenarios, face several issues.
First, most top-performing methods involve \textit{complex multi-stage training processes} to enhance prediction performance.
Methods such as \cite{DBLP:conf/eccv/YuanK20, DBLP:conf/iccv/MaoLS21} necessitate multiple training rounds for motion mode coverage and motion validity.
Recent diffusion model (DM)-based methods also require extra training stages for output post-processing \cite{DBLP:conf/aaai/WeiSLLLSH23} and motion encoding-decoding \cite{barquero2023belfusion}. 
The multiple training stages require laborious engineering efforts in model tuning, making them less appealing for many applications.
Second, stochastic HMP works often generate \textit{inconsistent or even unrealistic motion} with respect to the provided history \cite{DBLP:conf/eccv/YuanK20, DBLP:conf/mm/DangNLZL22}.
To regularize predictions and enhance diversity, these methods either incorporate explicit diversity-promoting losses \cite{DBLP:conf/iccv/MaoLS21} or construct additional sampling spaces \cite{DBLP:conf/mm/DangNLZL22} to avoid posterior collapse.
Such methods frequently result in sub-optimal predictions that deviate from the historical motion context.
This deviation can, at times, result in pose sequences that are entirely unrealistic from a physical standpoint.
Although recent works \cite{barquero2023belfusion} have begun to address this issue, ensuring the predicted motion that is both realistic and seamlessly synchronized with the provided motion history remains a significant challenge.

Intuitively, to mitigate these issues, it is reasonable for stochastic models to utilize the GCN-DCT design proven effective in deterministic contexts \cite{DBLP:conf/iccv/MaoLSL19, DBLP:conf/cvpr/MaNLZL22}, as their strong performance suggests a potential reduction in the cumbersome training pipelines and prediction inconsistency.
Surprisingly, this is not the case.
Most stochastic methods learn motion distributions through encoding-decoding in the latent space \cite{DBLP:conf/eccv/YuanK20, DBLP:conf/cvpr/SalzmannPR22, barquero2023belfusion, DBLP:conf/aaai/WeiSLLLSH23}, not preserving motion's spatial-temporal structure. 
This raises the question why such \textit{model design gap} exists between deterministic and stochastic HMP works.
To study this, we delve into the efficacy of the GCN-DCT design in deterministic models.
We find that these models excel primarily due to GCN's spatial modeling capability and, crucially, DCT's prowess in temporal modeling. 
As joint motion is temporally smooth, the DCT space significantly decreases the variability among elements between the last-observation-padded sequence and the groundtruth, as illustrated in \cref{fig:dct}, making it easier to learn than in the pose space. 
This ease of learning is evidenced by the universal use of global residual connections \cite{DBLP:conf/iccv/DangNLZL21} for residual learning across these methods \cite{DBLP:conf/iccv/MaoLSL19, DBLP:conf/eccv/MaoLS20, DBLP:journals/ijcv/MaoLSL21, DBLP:conf/eccv/CaiHWC0YLYZSLLM20}. 
However, this is not the case for stochastic models. 
The DM-based models primarily predict noise, and methods predicting motion equally find no learning benefits from concatenating motion history with noise, due to a much larger input-groundtruth discrepancy in the DCT space, a challenge also depicted in \cref{fig:dct}.
This fundamental difference explains why the GCN-DCT design has not been extensively adopted in stochastic HMP methods.

\begin{figure*}[!t]
\centering
\includegraphics[width=0.9\linewidth]{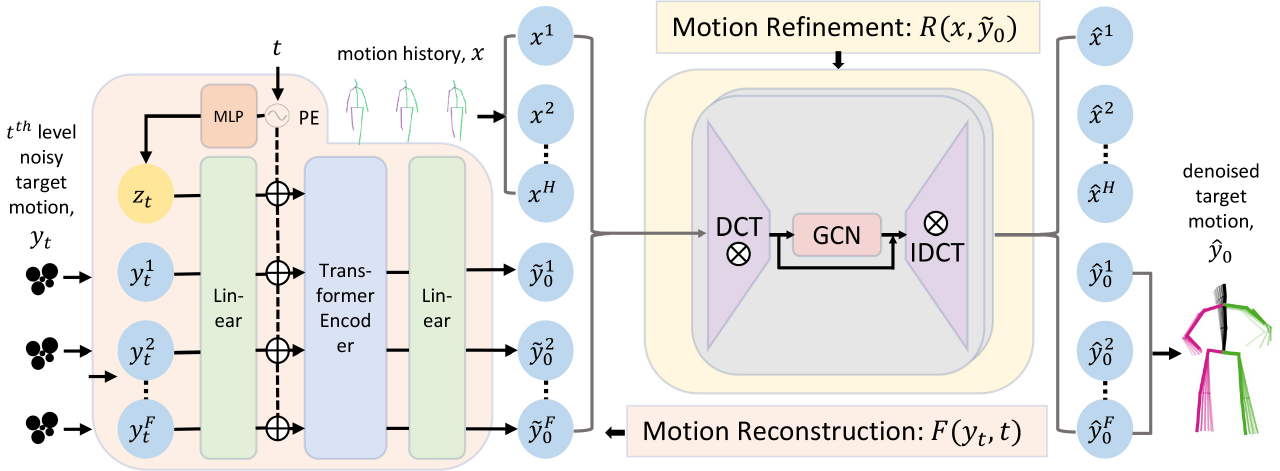}
\caption{
    Architecture of \name's predictor $G_{\theta}(\cdot)$. Inputs include the $t^{th}$ level target noisy motion $y_t$, motion history $x$, and time step $t$. The motion predictor operates in two stages: (1) the Transformer-based motion reconstruction module $F(\cdot)$ initially reconstructs $\tilde{y}_0$ from $y_t$ and $t$, and (2) the GCN-based motion refinement module $R(\cdot)$ then generates the complete motion sequence using the concatenated inputs of $x$ and $\tilde{y}_0$. IDCT stands for Inverse DCT and PE for Positional Encoding.}
\label{fig:mdl}
\end{figure*}

Building on this insight, we suggest that incorporating a pre-processing step to reconstruct smooth future motion from noise could simplify the learning process. 
By using sequences padded with the reconstructions as inputs for a GCN-DCT design, we can mirror the reduced learning difficulty observed in deterministic HMP works \cite{DBLP:conf/iccv/MaoLSL19, DBLP:conf/eccv/MaoLS20, DBLP:conf/cvpr/MaNLZL22}.
To this end, we introduce \name, a \textit{single-stage} DM-based framework tailored to \textit{consistent} HMP with its predictor architecture shown in \cref{fig:mdl}.
To generate a smooth future pose sequence, \name's motion predictor starts with a Transformer-based network for initial reconstruction of corrupted motion.
Then, a GCN is employed to refine the generated motion in the DCT space, using the concatenation of provided history and reconstructed sequence.
As such, \name~explicitly captures the spatio-temporal dependencies of human motion as a graph, which most stochastic HMP works have overlooked.
Importantly, \name~adopts a direct motion prediction strategy \cite{tevet2023human}, diverging from the common noise prediction scheme \cite{DBLP:conf/nips/HoJA20, barquero2023belfusion, chen2023humanmac}. 
This approach allows \name~to integrate a structure-aware loss that accounts for skeletal structure, further easing its learning process.
Moreover, with a simple yet effective adjustment to the standard cosine variance scheduler, we additionally elevate the accuracy and diversity of \name's generated motion samples.

Our contributions are summarized as follows. (1) We propose a single-stage, end-to-end diffusion framework for stochastic HMP, generating significantly more coherent and realistic motion than previous methods.
(2) We design a motion generator which combines Transformer and GCN to capture spatial-temporal dynamics of human motion in DCT space. To the best of our knowledge, \name~\\
represents the first exploration of integrating GCN-DCT design with DMs for stochastic HMP.
(3) We conduct comprehensive analyses to validate the efficacy of \name.
Benchmark results show that \name~outperforms previous approaches, achieving an improvement of at least 35\% in fidelity metrics, establishing it as a robust new baseline in the field.


\section{Related Work}
\label{sec::rela}

\subsubsection{Human Motion Prediction.}

Early efforts in HMP focused on deterministic settings \cite{DBLP:conf/iccv/FragkiadakiLFM15, DBLP:conf/cvpr/MartinezB017, DBLP:conf/cvpr/LiZLL18, DBLP:conf/3dim/AksanKCH21, DBLP:conf/ijcai/BouaziziHKDB22, DBLP:conf/wacv/GuoDSLAM23, DBLP:conf/cvpr/JainZSS16, DBLP:conf/cvpr/ButepageBKK17}, aiming to predict one most likely pose sequence. 
A key development in this domain was initiated by Mao \etal \cite{DBLP:conf/iccv/MaoLSL19}, which popularized modeling motion in DCT space using GCNs \cite{DBLP:conf/iccv/MaoLSL19, DBLP:conf/eccv/MaoLS20, DBLP:journals/ijcv/MaoLSL21, DBLP:conf/iccv/DangNLZL21, DBLP:conf/eccv/CaiHWC0YLYZSLLM20}. 
However, deterministic methods cannot model motion distributions and are thus not suitable for stochastic HMP.
To this end, generative methods \cite{xu22stars, DBLP:conf/cvpr/AliakbarianSSPG20, DBLP:conf/cvpr/GuoBAM22, DBLP:conf/cvpr/GuptaJFSA18, DBLP:conf/cvpr/LeeCVCTC17, DBLP:conf/cvpr/BarsoumKL18, DBLP:conf/aaai/KunduGB19, DBLP:conf/eccv/YuanK20} are proposed.
However, it is surprising that the top-performing stochastic works \cite{DBLP:conf/mm/DangNLZL22, DBLP:conf/cvpr/BlattmannMDO21a, DBLP:conf/cvpr/SalzmannPR22, barquero2023belfusion, posegpt} rarely utilize the GCN-DCT design which has been proven effective in deterministic settings. 
Instead, they opt for learning via encoding-decoding in the latent space, which does not preserve motion's spatial-temporal structure. 
While few works \cite{chen2023humanmac} attempt to explicitly exploit spatial-temporal patterns from a motion completion perspective, they, along with encoding-decoding methods, face issues such as complex, multi-stage training pipelines \cite{DBLP:conf/cvpr/Ma0HTC22, DBLP:conf/aaai/WeiSLLLSH23} and sub-optimal predictions that are often inconsistent with the provided history. 

Recently, few DM-based approaches \cite{DBLP:conf/aaai/WeiSLLLSH23, barquero2023belfusion, chen2023humanmac} are proposed due to their ability to produce more diverse, higher-quality samples compared to generative adversarial network (GAN) and variational autoencoder (VAE)-based methods.
Wei \etal proposed MotionDiff \cite{DBLP:conf/aaai/WeiSLLLSH23}, a two-stage framework that consists of a Transformer-based noise predictor for motion generation, and a pretrained network to enhance sample diversity as a post-processing step.
To address history-future inconsistency, Barquero \etal \cite{barquero2023belfusion} also took a two-stage approach to model the diffusion process in the behavioral space rather than the coordinate space.
Chen \etal proposed HumanMAC \cite{chen2023humanmac}, which uses a Transformer-based module with mask modeling to achieve single-stage learning.
Our \name~takes this one step further by exploring the potential of the GCN-DCT design with a pre-processing Transformer unit to model the motion denoising process. 
This synthesis allows \name~to capture the intricate spatial-temporal dynamics of human motion, and it ensures the efficiency of single-stage training coupled with unmatched performance in generating consistent, realistic prediction samples.

\subsubsection{Denoising Diffusion Models.}

Denoising diffusion probabilistic models (DD\-PMs) \cite{DBLP:conf/nips/GoodfellowPMXWOCB14, DBLP:conf/nips/HoJA20, DBLP:conf/icml/NicholD21, DBLP:conf/iclr/SongME21, DBLP:conf/iclr/WangZHCZ23, DBLP:conf/iclr/XiaoKV22} have recently received significant attention and have been applied in many fields \cite{DBLP:conf/nips/DhariwalN21, DBLP:journals/corr/abs-2209-04747, DBLP:conf/iclr/KongPHZC21, DBLP:conf/ijcai/HuangL0S00Z22, DBLP:conf/cvpr/WhangDTSDM22, DBLP:conf/iclr/PearceRKBSGMTMH23} due to their superior sample quality and diversity. 
DDPMs define a diffusion process in which random noise is gradually added to the data using a Markov chain, and then learn how to reverse the process to generate desired data samples. 
In the field of text-driven human motion synthesis, prior works such as MDM \cite{tevet2023human}, MotionDiffuse \cite{zhang2022motiondiffuse}, PhysDiff \cite{yuan2023physdiff} and MotionGPT \cite{jiang2023motiongpt} leverage DMs with Transformers to generate human motion via natural languages.
The advancements in DMs have also enabled works such as Diffusion-Conductor \cite{DBLP:journals/corr/abs-2306-10065}, EDGE \cite{DBLP:conf/cvpr/TsengCL23}, and MoFusion \cite{DBLP:conf/cvpr/DabralMGT23} in generating human motions synchronized with audio and music.

One bottleneck of DDPMs is their efficiency, as they typically require a large number of denoising steps to generate one sample.
Numerous efforts have aimed to tackle this issue, devising methods for fast sampling \cite{DBLP:conf/iclr/ZhangC23, DBLP:conf/nips/0011ZB0L022, DBLP:journals/corr/abs-2211-01095} and reducing the resolution of data \cite{DBLP:conf/cvpr/RombachBLEO22} through auto-encoding.
Though not in the thousands, HMP works such as \cite{DBLP:conf/aaai/WeiSLLLSH23, chen2023humanmac} still require a large number of denoising steps even when using advanced samplers \cite{DBLP:conf/iclr/SongME21}.
Benefiting from learning motion directly instead of noise with a motion predictor that reduces learning difficulty through the GCN-DCT design, \name~achieves state-of-the-art performance in both prediction accuracy and fidelity with only a few diffusion steps, without the need for fast sampling techniques, while maintaining an appropriate level of diversity.
\section{Methodology}
\label{sec::method}

\subsection{Problem Definition}

Given a motion history $x^{1:H} = \{x^i\}_{i=1}^H$ of length $H$, our objective is to predict the subsequent $F$ poses $x^{H+1:H+F} = \{x^i\}_{i=H+1}^{H+F}$.
In stochastic HMP, we forecast multiple pose sequences from a single motion history, denoting one predicted sequence as $y^{1:F} \coloneqq x^{H+1:H+F}$.
Each pose at time step $i$ is represented as $x^i \in \sR^{J \times 3}$ where $J$ being the number of body joints. 
Superscripts in $x^{1:H}$ and $y^{1:F}$ may be omitted when contextually clear.

\subsection{Conditional Motion Diffusion}
\label{sec::cmd}

Let $\{y_t\}_{t=0}^T$ denote a general Markov noising process where $y_0$ represents the true data samples.
The forward unconditional diffusion transitions are denoted as:
\begin{equation}
\label{eq::forward_diff}
    q(y_t|y_{t-1})=\mathcal{N}(y_t; \sqrt{\alpha_t} y_{t-1},(1-\alpha_t) \textrm{I}),
\end{equation}
where $\{\alpha_t\}_{t=0}^T \in [0,1]$ control the noise level, and can either be fixed \cite{DBLP:conf/nips/HoJA20} or learned \cite{DBLP:journals/corr/abs-2107-00630}.
To estimate the true data distribution, the reverse diffusion process is constructed to progressively denoise the corrupted data samples $y_t$ from $t=T$ to $t=1$ as:
\begin{equation}
\label{eq::reverse_diff}
    p_{\theta}(y_{t-1}|y_t)=\mathcal{N}(y_{t-1}; \mu_{\theta}(y_t, t),\sigma_{\theta}^2(y_t, t)\textrm{I}).
\end{equation}
In our HMP context, we need to extend the above formulation to the conditional case. Specifically, the reverse diffusion transition in \cref{eq::reverse_diff} becomes:
\begin{equation}
\label{eq::reverse_hmp}
    p_{\theta}(y_{t-1}|y_t, x)=\mathcal{N}(y_{t-1}; \mu_{\theta}(y_t, x, t),\sigma_{\theta}^2(y_t, x, t)\textrm{I}),
\end{equation}
where $x$ represents the motion history and $y_t$ represents the $t^{th}$ level target noisy motion.
In the seminal work \cite{DBLP:conf/nips/HoJA20}, Ho \etal proposed to (1) fix $\sigma_{\theta}^2(\cdot)$, (2) predict noise $\epsilon$ using a noise predictor $\epsilon_\theta(\cdot)$ instead of $\mu$ by reparameterization, and (3) optimize the model with objective $L(\theta)=\mathbb{E}_t[\|\epsilon-\epsilon_\theta(\cdot)\|^2]$.
These techniques are quickly adopted by later DM-based works \cite{DBLP:conf/aaai/WeiSLLLSH23, chen2023humanmac} due to the simplicity and great performance they offer.

\subsection{Motion Diffusion Pipeline}
\label{sec::pipeline}

\subsubsection{Prediction Target.}
However, the $\epsilon$-prediction target in the diffusion formulation presented above hinders HMP models from enjoying the aforementioned GCN-DCT design's learning benefits, and also prevents them from leveraging various motion losses that have been extensively studied in previous works \cite{DBLP:conf/cvpr/KocabasAB20, DBLP:journals/tog/HarveyYNP20}.
To address these, we choose to predict the future motion directly \cite{tevet2023human}.
Specifically, instead of predicting $\epsilon$, we choose another reparameterization such that $\hat{y}_0 \leftarrow G_{\theta}(y_t, x, t)$, where $\hat{y}_0$ is the learned approximation of target future motion $y_0$.
This predicted motion $\hat{y}_0$ is then diffused back through $t-1$ steps and, together with the provided motion history $x$, are used to generate the subsequent predictions in the sampling chain. 
With $\bar{\alpha}_{t} = \prod_{s=1}^{t} \alpha_s$, the forward diffusion process in \cref{eq::forward_diff} can be simplified as:
\begin{equation}
\label{eq::forward_simple}
    q(y_{t}|y_0)=\mathcal{N}(y_{t}; \sqrt{\bar{\alpha}_{t}} y_0,(1-\bar{\alpha}_{t}) \textrm{I}).
\end{equation}

\subsubsection{Generator Architecture.}

With the $y_0$-prediction objective, our motion generator can benefit from the effective GCN-DCT design used in deterministic HMP works.
In particular, $G_{\theta}(\cdot)$ leverages the spatio-temporal graph structure of motion data.
Our designed network, shown in \cref{fig:mdl}, consists of (1) a Transformer-based reconstruction module $F(\cdot)$ and (2) a GCN-based refinement module $R(\cdot)$.

First, we utilize a Transformer-based module, denoted as $F(y_t, t)$, to generate an ``initial reconstruction'' $\Tilde{y}_0$ from the target noisy motion $y_t$, without considering the motion history.
$F(y_t, t)$ explicitly models temporal correlations across noisy frames using a Transformer encoder.
The time step $t$ is first mapped to an embedding and then projected to the same latent dimension as the noisy motion frames $\{y_t^i\}_{i=1}^F$ via a feedforward network, producing a time token $z_t$.
The noisy motion frames $\{y_t^i\}_{i=1}^F$ are first projected and then summed with positional encodings \cite{DBLP:conf/nips/VaswaniSPUJGKP17} to obtain positional information.
These transformed motion frames are then prepended with the time token $z_t$ and fed into the Transformer encoder.
To derive the initial reconstruction $\Tilde{y}_0$ from $F(\cdot)$, we discard the first output token which corresponds to $t$, and subsequently project the remaining learned representations back into the pose dimension of $J \times 3$.
Using $F(\cdot)$ as a pre-processing step is crucial as it yields a smoother motion representation in the coordinate space compared to $y_t$, particularly in the early phases of denoising. 
The learned representation $\Tilde{y}_0$ can be seamlessly integrated with the given motion history $x$, simplifying the learning process for the refinement module. 
The effectiveness of $F(\cdot)$ is further demonstrated in \cref{subsec::ablation}.

For the refinement module $R(x, \Tilde{y}_0)$, we adopt the GCN-based architecture \cite{DBLP:journals/corr/abs-2305-04443} from deterministic HMP research due to its effectiveness. 
The module $R(\cdot)$ begins by concatenating the inputs $x$ and $\Tilde{y}_0$. 
It then progressively refines the entire motion trajectory by alternating between the pose space and frequency space using DCT and its inverse (IDCT). 
The process starts by converting the motion trajectory into DCT coefficients, which are then processed by multiple GCN layers to capture the spatial-temporal relationships among joints. 
Next, the refined motion representation is converted back to the pose space and is used as the input for the next GCN block. 
The final predicted future motion $\hat{y}_0$ is obtained by removing the segment that corresponds to the known motion history.
That is:
\begin{equation}
\label{eq::network}
    \hat{y}_0 \leftarrow [\hat{x}; \hat{y}_0] = R(x, F(y_t, t)) \coloneqq G_{\theta}(y_t, x, t).
\end{equation}

\subsubsection{Variance Scheduler.}
The variance scheduler $\{1 - \alpha_t\}_{t=0}^T$ is essential for the performance of DMs.
The linear scheduler \cite{DBLP:conf/nips/HoJA20} and the cosine scheduler \cite{DBLP:conf/icml/NicholD21} are commonly used due to their simplicity and performance they provide.
However, the suitability of these schedulers for specific cases, such as ours, warrants further investigation.
As mentioned, our generator $G_{\theta}(y_t, x, t)$ aims to (1) directly predict future motion and (2) incorporate the historical motion sequence $x^{1:H}$ at each step of denoising. 
This approach differs markedly from the multimodal conditional settings found in other DM applications \cite{yuan2023physdiff, DBLP:journals/tog/AlexandersonNBH23}, where the temporal behavioral guidance is not as explicitly defined.

To this end, we study how standard schedulers are designed.
\Cref{eq::forward_simple} is equivalent to:
\begin{equation}
\label{eq::forward_rep}
    y_t=\sqrt{\bar{\alpha}_t} y_0 + \sqrt{1-\bar{\alpha}_t}\epsilon, 
\end{equation}
where $\epsilon \sim \gN(0, \textrm{I})$.
In the theoretical setup, $\bar{\alpha}_0 = 1$ and $\bar{\alpha}_T = 0$, effectively transforming clean data into standard Gaussian noise.
Following this premise, both linear and cosine schedulers are empirically designed to have $\bar{\alpha}_0 \approx 1$.

However, our empirical observations (\cref{tab:abla_dm}) suggest these standard schedulers, while effective in many scenarios, do not facilitate the diversity and accuracy of samples necessary for our model to excel.
We hypothesize that having $\bar{\alpha}_0$ close to $1$ prevents $G_{\theta}(y_t, x, t)$ from producing accurate and diverse samples.
Specifically, since we feed the clean motion history directly into $G_{\theta}(\cdot)$ and aim to predict $y_0$ by modeling $x^{1:H}$ explicitly, the strong guidance from $x^{1:H}$ and the strong spatial-temporal modeling of $G_{\theta}(\cdot)$ make the prediction task overly simple as the reverse process progresses.
Due to the poor mode coverage of the current HMP datasets with their limited size, predictions may not be multimodal, resulting in sub-optimal sample diversity and accuracy.

To address the issue, rather than developing complex methods to promote sample diversity, we simply modify the original cosine scheduler to better suit our HMP framework.
Specifically, we have:
\begin{equation}
\label{eq::scheduler}
    \bar{\alpha}_t = \cos \left(\frac{t / T+ 1}{2} \cdot \frac{\pi}{2}\right)^2.
\end{equation}
By setting the offset to $1$ as opposed to the commonly used $0.008$ and relax the $\bar{\alpha}_0 \approx 1$ restriction, we establish an initial value of $\bar{\alpha}_0 = \cos(\pi/4)^2 = 0.5$. 
This change ensures that the task of predicting $y_0$ remains non-trivial even at the final denoising stages when $t$ is close to $0$, and consequently, obliges the model to consistently tackle the prediction task throughout the entire diffusion process, without an over-reliance on the clarity of the motion history guidance. 

\subsection{Learning Algorithm}
\label{sec::learning}

As outlined in \cref{sec::intro}, aiming to predict motion directly enables us to utilize geometric losses to supervise the model $G_{\theta}(\cdot)$.
To this end, we utilize a structure-aware loss \cite{DBLP:journals/corr/abs-2305-04443} to take into account the details of the structure of human motion. 
The structure-aware loss weights all joints differently to reflect their relative importance in the motion context.
For a single motion trajectory, the loss function for \name~is expressed as:
\begin{equation}
\label{eq::exp_loss}
\gL_{\theta}(G, y_0, x)=\E_{\substack{y_0 \sim q( \cdot \mid x) \\t \sim[1, T]}}\gL_{\textrm{rec}}(G_{\theta}(y_t, x, t), y_0, x). 
\end{equation}
The reconstruction loss is defined by:
\begin{equation}
\label{eq::total_loss}
\gL_{\textrm{rec}}=\frac{1}{J}\sum_{j=1}^J(\gamma \cdot \|(x^j-\hat{x}^j) \cdot \lambda^j\|_1 + \|(y_0^j-\hat{y}_0^j) \cdot \lambda^j\|_1),
\end{equation}
where the superscript $j$ indicates the joint index, $\lambda^j$ is the weight assigned to each joint, and $\gamma$ is a hyperparameter balancing the importances of the reconstruction of motion history and the prediction of future.
Importantly, \name~not only predicts $y_0$ but also reconstructs $x$, ensuring a global awareness of the entire motion trajectory and thus enhancing prediction accuracy.
The weights for each joint are determined based on the kinematic structure of the human body, prioritizing joints prone to more dynamic movements, and do not require learning.
For weight $\lambda^j$ derivation details, please refer to the supplementary material.

Additionally, we use the relaxation technique \cite{DBLP:conf/iccv/MaoLS21, barquero2023belfusion} to further promote the prediction diversity.
For each motion history, we generate $k$ target motion trajectories and only optimize $\gL_{\theta}$ towards the most accurate prediction.
That is:
\begin{equation}
\label{eq::loss_relax}
\gL_{\textrm{final}}=\min_k\gL_{\theta}(G^k, y_0, x),
\end{equation}
where $G^k$ is the $k^{th}$ generated motion trajectory based on the original pose sequence $[x; y_0]$.

\section{Experiments}
\label{sec::exp}

\begin{table*}[t]\footnotesize	
\centering
\caption{Quantitative results for Human3.6M dataset \cite{DBLP:journals/pami/IonescuPOS14}. The best results are highlighted in \textbf{bold}. The symbol `-' indicates that the results are not reported in the baseline work. For all metrics except for APD, lower is better.} 
\resizebox{\linewidth}{!}{
\begin{tabular}{c | c c | c c c c c c c c}
\hline \multicolumn{1}{c}{Type} & Method & \multicolumn{1}{c}{One-Stage} & APD $\uparrow$ & APDE $\downarrow$ & ADE $\downarrow$ & FDE $\downarrow$ & MMADE $\downarrow$ & MMFDE $\downarrow$ & CMD $\downarrow$ & FID $\downarrow$\\

\hline 
\multirow{2}{*}{GAN-based}
& HP-GAN \cite{DBLP:conf/cvpr/BarsoumKL18} & \cmark & 7.214 & - & 0.858 & 0.867 & 0.847 & 0.858 & - & - \\

& DeLiGAN \cite{DBLP:conf/cvpr/GurumurthySB17} & \cmark & 6.509 & - & 0.483 & 0.534 & 0.520 & 0.545 & - & - \\
\hline
\multirow{9}{*}{VAE-based}
& TPK \cite{DBLP:conf/iccv/WalkerMGH17} & \cmark & 6.723 & 1.906 & 0.461 & 0.560 & 0.522 & 0.569 & 6.326 & 0.538 \\

& Motron \cite{DBLP:conf/cvpr/SalzmannPR22} & \cmark & 7.168 & 2.583 & 0.375 & 0.488 & 0.509 & 0.539 & 40.796 & 13.743 \\

& DSF \cite{DBLP:conf/iclr/0007K20} & \xmark & 9.330 & - & 0.493 & 0.592 & 0.550 & 0.599 & - & - \\
& DLow \cite{DBLP:conf/eccv/YuanK20} & \xmark & 11.741 & 3.781 & 0.425 & 0.518 & 0.495 & 0.531 & {4.927} & 1.255 \\

& GSPS \cite{DBLP:conf/iccv/MaoLS21} & \xmark & {14.757} & 6.749 & 0.389 & 0.496 & 0.476 & 0.525 & 10.758 & 2.103 \\

& DivSamp \cite{DBLP:conf/mm/DangNLZL22} & \xmark & 15.310 & 7.479 & {0.370} & 0.485 & {0.475} & {0.516} & 11.692 & 2.083 \\
\hline
\multirow{4}{*}{DM-based}
& MotionDiff \cite{DBLP:conf/aaai/WeiSLLLSH23} & \xmark & \textbf{15.353} & - & 0.411 & 0.509 & 0.508 & 0.536 & - & - \\
& HumanMAC \cite{chen2023humanmac} & \cmark & 6.301 & - & 0.369 & 0.480 & 0.509 & 0.545 & - & - \\
& BeLFusion \cite{barquero2023belfusion} & \xmark & 7.602 & {1.662} & 0.372 & {0.474} & \textbf{0.473} & 0.507 & 5.988 & {0.209} \\
\rowcolor{gray}
\cellcolor{white} & Ours & \cmark & 7.632 & \textbf{1.609} & \textbf{0.350} & \textbf{0.458} & 0.494 & \textbf{0.506} & \textbf{3.202} & \textbf{0.102} \\
\hline
\end{tabular}}
\label{tab:quan_h36m}
\end{table*}

\subsection{Datasets}
\label{subsec::dataset}

\noindent\textbf{Human3.6M} \cite{DBLP:journals/pami/IonescuPOS14}, the most widely used dataset for stochastic HMP, contains motion clips of 7 subjects performing 15 different actions recorded at 50 Hz.
For a fair comparison with previous works, we adopt the evaluation protocol of \cite{barquero2023belfusion}, where a 16-joint skeleton is used for human structure modeling.
We predict 2s (100 frames) based on 0.5s (25 frames) of observation.

\noindent\textbf{AMASS} \cite{DBLP:conf/iccv/MahmoodGTPB19} unifies multiple Mocap datasets, such as HumanEva-I \cite{DBLP:journals/ijcv/SigalBB10} using a shared SMPL \cite{DBLP:journals/tog/LoperM0PB15} parameterization for human skeleton modeling.
As a multi-dataset collection, AMASS can be used to perform cross-dataset evaluation to examine a model's generalization ability.
Evaluation settings such as frame rate and dataset partition are all set to be the same as in previous works \cite{barquero2023belfusion} for fair comparisons.
We predict 2s (120 frames) into the future based on 0.5s (30 frames) of observation after downsampling.

\subsection{Experimental Setup}
\label{subsec::setup}

\subsubsection{Evaluation Metrics.}
Following previous work \cite{barquero2023belfusion}, we use a comprehensive set of metrics to evaluate \name~quantitatively.
(1) Average Pairwise Distance (APD) computes the averaged $\ell_2$ distance between all generated sample pairs to measure sample diversity. 
(2) The Average and (3) the Final Displacement Errors (ADE and FDE) calculate the averaged all-time and the last-frame $\ell_2$ distances respectively between the groundtruth and the closest prediction, measuring sample accuracy.
The multimodal versions of ADE and FDE, (4) MMADE, and (5) MMFDE assess a method’s ability to produce multimodal predictions, whose groundtruth is obtained by grouping similar observations.
To quantify the realism of motion, (6) the Fr\'echet Inception Distance (FID) is used to assess the similarity between the distributions of generated and real motions.

Recently, Barquero \etal \cite{barquero2023belfusion} proposed two metrics to better quantify a model's ability to produce behaviorally consistent motion. 
(6) The area of the Cumulative Motion Distribution (CMD) measures the difference between the areas under the cumulative true motion and predicted motion distributions, capturing the plausibility of predicted motion at a global level.
To analyze to what extent the diversity is properly modeled, (7) the Average Pairwise Distance Error (APDE) is defined as the absolute error between the APD of the multimodal groundtruth and the APD of the predictions.
For metric calculation details, please refer to the supplementary material.

\begin{table*}[t]\footnotesize	
\centering
\caption{Quantitative results for AMASS dataset \cite{DBLP:conf/iccv/MahmoodGTPB19}. The best results are highlighted in \textbf{bold}. The symbol `-' indicates that the results are not reported in the baseline work. As AMASS does not contain class labels, the FID metric is not used for evaluation.} 
\resizebox{\linewidth}{!}{
\begin{tabular}{c | c c | c c c c c c c c}
\hline \multicolumn{1}{c}{Type} & Method & \multicolumn{1}{c}{One-Stage} & APD $\uparrow$ & APDE $\downarrow$ & ADE $\downarrow$ & FDE $\downarrow$ & MMADE $\downarrow$ & MMFDE $\downarrow$ & CMD $\downarrow$\\

\hline
\multirow{4}{*}{VAE-based}
& TPK \cite{DBLP:conf/iccv/WalkerMGH17} & \cmark & 9.283 & 2.265 & 0.656 & 0.675 & 0.658 & 0.674 & 17.127 \\
& DLow \cite{DBLP:conf/eccv/YuanK20} & \xmark & {13.170} & 4.243 & 0.590 & 0.612 & 0.618 & 0.617 & {15.185} \\
& GSPS \cite{DBLP:conf/iccv/MaoLS21} & \xmark & 12.465 & 4.678 & 0.563 & 0.613 & 0.609 & 0.633 & 18.404 \\
& DivSamp \cite{DBLP:conf/mm/DangNLZL22} & \xmark & \textbf{24.724} & 15.837 & 0.564 & 0.647 & 0.623 & 0.667 & 50.239 \\
\hline
\multirow{3}{*}{DM-based}
& HumanMAC \cite{chen2023humanmac} & \cmark & 9.321 & - & 0.511 & 0.554 & 0.593 & 0.591 & - \\
& BeLFusion \cite{barquero2023belfusion} & \xmark & 9.376 & \textbf{1.977} & {0.513} & {0.560} & {0.569} & {0.585} & 16.995 \\
\rowcolor{gray}
\cellcolor{white} & Ours & \cmark & 10.848 & 2.328 & \textbf{0.494} & \textbf{0.547} & \textbf{0.469} & \textbf{0.466} & \textbf{9.636}
\\
\hline
\end{tabular}}
\label{tab:quan_amass}
\end{table*}

\subsubsection{Baselines.}
For Human3.6M quantitative evaluation, \name~is compared with GAN-based approaches HP-GAN \cite{DBLP:conf/cvpr/BarsoumKL18}, DeLiGAN \cite{DBLP:conf/cvpr/GurumurthySB17}, VAE-based methods TPK \cite{DBLP:conf/iccv/WalkerMGH17}, Motron \cite{DBLP:conf/cvpr/SalzmannPR22}, DSF \cite{DBLP:conf/iclr/0007K20}, DLow \cite{DBLP:conf/eccv/YuanK20}, GSPS \cite{DBLP:conf/iccv/MaoLS21}, DivSamp \cite{DBLP:conf/mm/DangNLZL22}, and DM-based methods MotionDiff \cite{DBLP:conf/aaai/WeiSLLLSH23}, HumanMAC \cite{chen2023humanmac}, BeLFusion \cite{barquero2023belfusion}.
A selection of these, representing the most competitive methods, is further evaluated quantitatively on the AMASS dataset. 
For qualitative analysis, we compare \name~with DLow, GSPS, DivSamp, and BeLFusion.

\subsubsection{Implementation Details.}

\begin{wrapfigure}[14]{R}{0.5\textwidth}
\centering
\includegraphics[width=0.5\textwidth]{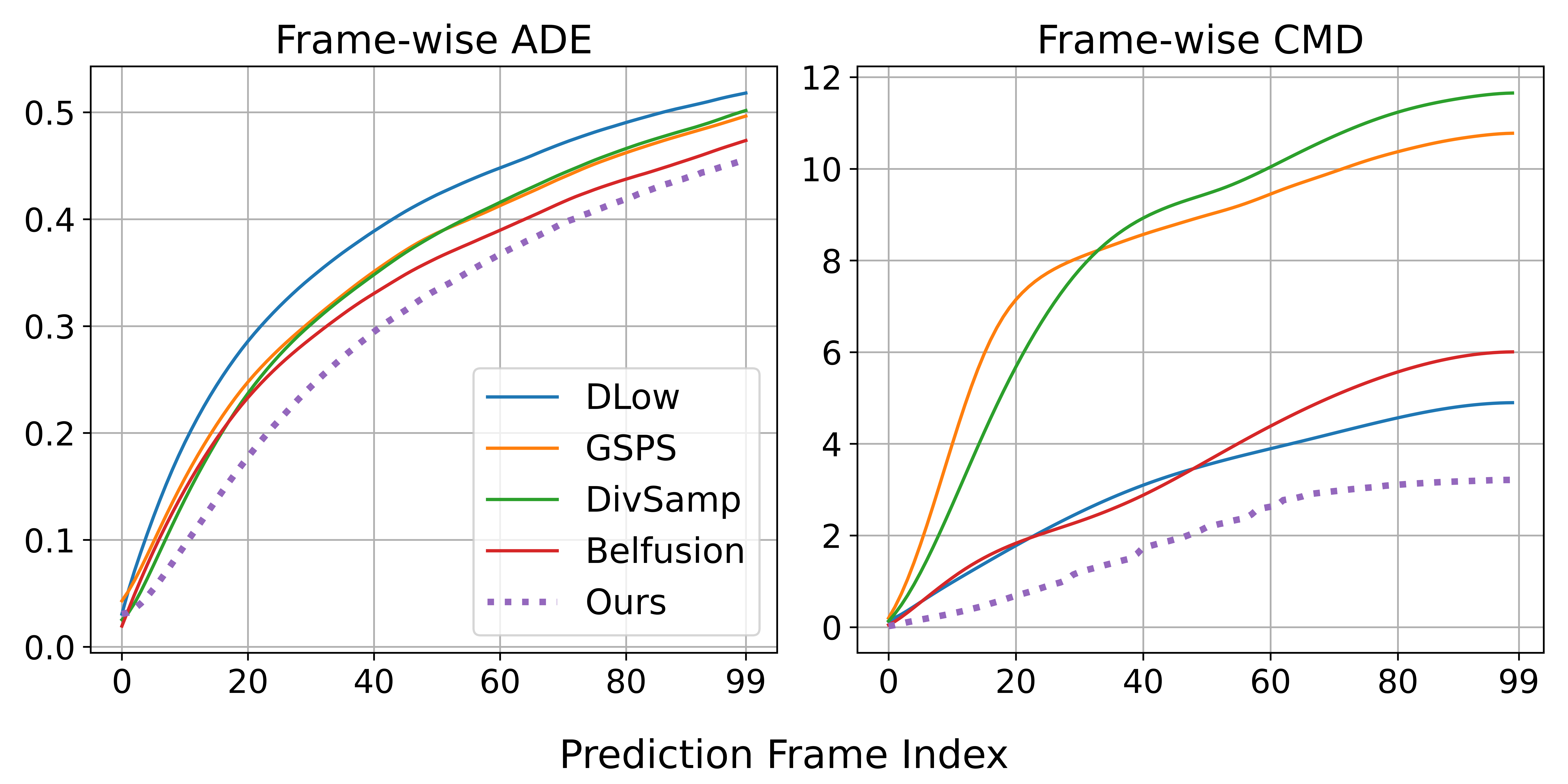}
\caption{Left: ADE computed at each prediction frame of state-of-the-art methods. Right: CMD computed up to each prediction frame. Both experiments are conducted on Human3.6M dataset.}
\label{fig:per-step}
\end{wrapfigure}

We train \name~as a 10-step DM with standard DDPM \cite{DBLP:conf/nips/HoJA20} sampling.
We implement $F(y_t, t)$ using 8 Transformer encoder layers with a latent dimension of 512.
We use a 2-layer MLP to project the time step embedding to the transformer dimension.
The GCN-based refinement module $R(x, \Tilde{y}_0)$ consists of 3 blocks, each of which contains 2 GCN layers with a latent dimension of 256.
The latent dimension and the dropout ratio of the refinement module are 256 and 0.5.
Adam~\cite{DBLP:journals/corr/KingmaB14} is used for all experiments with 0.0001 as the initial learning rate.
For both datasets, \name~is trained for 500 epochs, and the learning rate starts to decay after the 200$^{th}$ epoch.
More implementation details can be found in the supplementary material.

\subsection{Results Compared with State of the Art}
\label{subsec::result}

\subsubsection{Quantitative Results.}

The main quantitative comparison results are shown in \cref{tab:quan_h36m,tab:quan_amass}.
For Human3.6M, we observe in \cref{tab:quan_h36m} that \name~outperforms previous methods on the accuracy metrics (ADE and FDE) by large margins, which underlines the plausibility of our predicted motions.
More notably, our method excels in generating behaviorally consistent and realistic future motions, evidenced by substantial improvements of $\mathbf{35\%}$ in CMD and $\mathbf{51\%}$ in FID over previous state of the art.
While our model does not achieve the highest scores in the diversity metric (APD), it shows the best performance in APDE. 
This demonstrates \name's capability to properly model the stochasticity of future motion based on the past. 
Furthermore, the frame-wise ADE and CMD results shown in \cref{fig:per-step} indicate that \name~can consistently outperform previous methods at each prediction frame.
For AMASS results in \cref{tab:quan_amass}, we achieve a competitive performance in APDE, and obtain state-of-the-art results in all other metrics (a $\mathbf{43\%}$ improvement on CMD) except APD.
This indicates again that \name~can generate consistent, realistic motion with proper diversity. 

\begin{figure*}[t]
\centering
\includegraphics[width=1.0\linewidth]{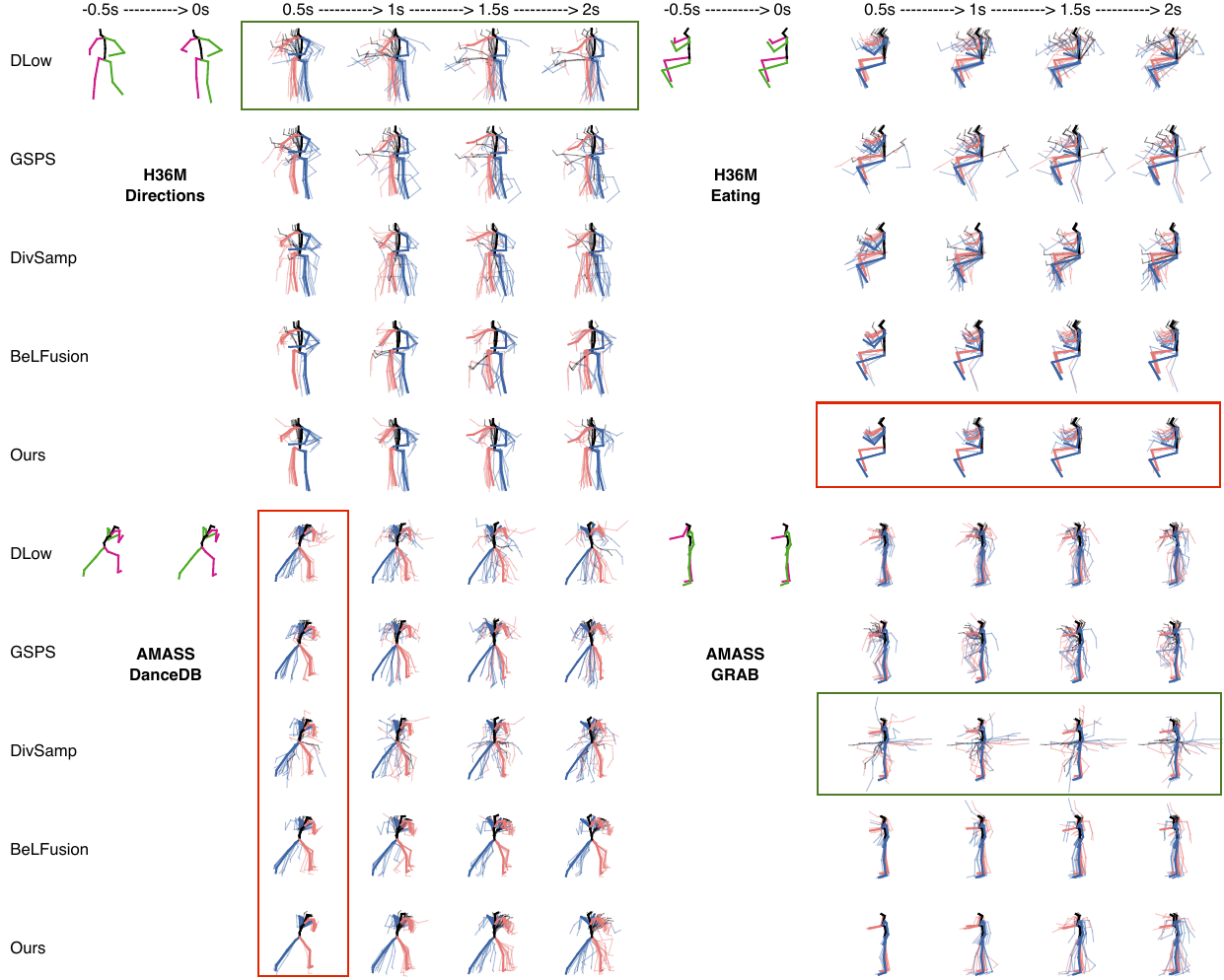}
\caption{Qualitative results of \name~compared with baseline methods. 
The upper block of rows corresponds to results obtained from the Human3.6M dataset, while the lower block of rows represents results from the AMASS dataset. 
The \textit{green-purple} and the \textit{blue-orange} skeletons denote the observed history and the predictions respectively.}
\label{fig:qual}
\end{figure*}
\subsubsection{Qualitative Results.}

In \cref{fig:qual}, we compare \name~against multiple state-of-the-art methods qualitatively on both datasets, superimposing 10 predictions beneath the groundtruth motion at each prediction frame. 
Two actions, \texttt{Directions} and \texttt{Eating}, from Human3.6M, and two sub-datasets, \texttt{DanceDB} and \texttt{GRAB} from AMASS, are showcased for this comparison.

The visualizations first confirm that \name~is capable of generating natural and coherent stochastic predictions that are well-aligned with the motion history, as our predicted motions are qualitatively more reasonable and contain fewer anomalies.
For instance, take the predictions at 0.5s for \texttt{DanceDB} from AMASS, highlighted in the lower left red box, where the initial poses predicted by \name~are closely aligned with the groundtruth and then gradually depict variations over time. 
This is in stark contrast to other baselines, which often exhibit sudden motion discontinuities.
Second, \name~demonstrates its ability to produce diverse predictions that are adapted to the context. 
For example, in the \texttt{Eating} action of Human3.6M (upper right red box), the predicted motions by \name~display various arm movements while maintaining the legs in a stationary position in most cases, signifying a realistic portrayal of the \texttt{Eating} action.
More importantly, \name~tends to generate much fewer unreasonable poses when compared to other methods. 
For instance, in the \texttt{Directions} action for DLow (upper left green box), we observe an unnatural sudden bend from many predictions, and in the \texttt{GRAB} action for DivSamp (lower right green box), many predicted poses start to float in the air, violating real-world physical rules. 
These comparative insights highlight the advantage of \name~in producing more plausible and contextually appropriate predictions than previously established methods.
more examples are provided in the supplementary material.

\subsection{Ablation Study}
\label{subsec::ablation}

In this section, we conduct an ablation analysis on the Human3.6M dataset to investigate how different design choices of \name~affect its motion modeling ability.
Additional studies are included in the supplementary material.

\begin{table}[t]\scriptsize
\centering
\caption{Ablation on \name's general architecture. In the \texttt{\textit{Sched.}} column, \cmark~denotes use of our proposed scheduler.} 
\resizebox{\linewidth}{!}{
\begin{tabular}{c c c | c c c c c c c c}
\hline $F(y_t, t)$ & $R(x, \Tilde{y}_0)$ & \multicolumn{1}{c}{\texttt{\textit{Sched.}}} & APD $\uparrow$ & APDE $\downarrow$ & ADE $\downarrow$ & FDE $\downarrow$ & MMADE $\downarrow$ & MMFDE $\downarrow$ & CMD $\downarrow$ & FID $\downarrow$\\
\hline 
\xmark & \xmark & \xmark & \textbf{12.880} & 4.854 & 0.959 & 1.000 & 0.987 & 1.004 & 966.716 & 1.047 \\
\cmark & \xmark & \xmark & 3.727 & 4.441 & 0.502 & 0.669 & 0.632 & 0.731 & \textbf{3.176 }& 0.167	 \\
\xmark & \cmark & \xmark & 6.858 & 1.835 & 0.539 & 0.678 & 0.625 & 0.694 & 197.105 & 0.474 \\
\cmark & \cmark & \xmark & 7.602 & \textbf{1.446} & 0.382 & 0.489 & 0.521 & 0.537 & 3.323 & 0.282 \\
\rowcolor{gray}
\cmark & \cmark & \cmark & 7.632 & 1.609 & \textbf{0.350} & \textbf{0.458} & \textbf{0.494} & \textbf{0.506} & 3.202 & \textbf{0.102} \\
\hline
\end{tabular}}
\label{tab:abla_gene}
\end{table}

\subsubsection{Framework Components.}
\begin{wrapfigure}[13]{r}{0.5\textwidth}
\centering
\includegraphics[width=0.5\textwidth]{figs/new_vis_recon_2.png}
\caption{Left: $y_T$, a Gaussian trajectory with $F = 100$ frames.
Right: $F(y_T, T)$, the reconstructed trajectory. 
Compared with $y_T$, $F(y_T, T)$ depicts a much smoother temporal pattern with lower variance.}
\label{fig:recon}
\end{wrapfigure}
We first evaluate the individual contributions of \name's components to its performance.
We test five variants of our framework by selectively disabling (1) the reconstruction module $F(\cdot)$, (2) the refinement module $R(\cdot)$, and (3) the proposed scheduler. 
If both modules are disabled, a plain GCN \cite{DBLP:conf/iccv/MaoLSL19} is used.
The results in \cref{tab:abla_gene} show that omitting the reconstruction module $F(\cdot)$ consistently leads to inferior performance, evidenced by higher ADE and FDE for accuracy and increased FID and CMD, indicating less realistic predictions compared to other variants.
This is concurrently supported by \cref{fig:recon}, where $\Tilde{y}_0$ produced by $F(\cdot)$ exhibits a much smoother temporal pattern than the target noisy motion $y_t$, thereby easing the subsequent task for the refinement module. 
Furthermore, the refinement module $R(\cdot)$ considerably improves both prediction accuracy and fidelity, benefits that are further enhanced by the proposed variance scheduler. 
These results demonstrate \name's effective use of the GCN-DCT design from deterministic works.

\begin{figure}[t]
\centering
\includegraphics[width=0.9\linewidth]{figs/diff_steps.png}
\caption{Ablation results on the number of diffusion steps. The bottom rightmost sub-figure shows the per-sample time spent in seconds on Human3.6M inference.}
\label{fig:abla_diff}
\end{figure}
\begin{table}[t]\scriptsize
\centering
\caption{\textbf{Left (a)}: Ablation on prediction target. \textbf{Right (b)}: Ablation on variance scheduler. Linear scheduler's results are not included as it causes \name~to diverge.}
\resizebox{\linewidth}{!}{
\begin{tabular}{c | c c c c c}
\hline \multicolumn{1}{c}{Target} & APD $\uparrow$ & ADE $\downarrow$ & MMADE $\downarrow$ & CMD $\downarrow$ & FID $\downarrow$ \\
\hline 
$\epsilon$ & \textbf{8.266} & 0.431 & 0.515 & 25.968 & 0.290 \\
\rowcolor{gray}
 $y_0$ (ours) & 7.632 & \textbf{0.350} & \textbf{0.494} & \textbf{3.202} & \textbf{0.102} \\
\hline
\end{tabular}
\begin{tabular}{c | c c c c c}
\hline\multicolumn{1}{c}{Scheduler} & APD $\uparrow$ & ADE $\downarrow$ & MMADE $\downarrow$ & CMD $\downarrow$ & FID $\downarrow$ \\
\hline 
Cosine & 7.602 & 0.382 & 0.521 & 3.323 & 0.282 \\
Sqrt. & 6.988 & 0.359 & 0.503 & \textbf{2.832} & 0.128 \\
\rowcolor{gray}
ours & 7.632 & \textbf{0.350} & \textbf{0.494} & 3.202 & \textbf{0.102} \\
\hline
\end{tabular}
}
\label{tab:abla_dm}
\end{table}

\subsubsection{Diffusion Model Setups.}

\begin{wrapfigure}[6]{R}{0.5\textwidth}\scriptsize	
\centering
\makeatletter\def\@captype{table}\makeatother
\caption{Effect of $\bar{\alpha}_0$.} 
\begin{tabular}{ccccccc}
\hline 
\multicolumn{1}{c}{$\bar{\alpha}_0$} & $\approx 1$ & 0.9 & 0.8 & 0.7 & 0.6 & \cellcolor{gray} 0.5 (ours) \\
\hline 
 ADE $\downarrow$ & 0.407 & 0.368 & 0.361 & 0.356 & 0.354 & \cellcolor{gray} \textbf{0.350} \\
 FID $\downarrow$ & 0.323 & 0.138 & 0.123 & 0.109 & 0.110 & \cellcolor{gray} \textbf{0.102} \\
\hline
\end{tabular}
\label{tab:alpha_effect}
\end{wrapfigure}

For DM setup, we study the impacts of (1) the choice of prediction target, (2) the choice of variance scheduler and (3) the number of denoising steps $T$.
The results are summarized in \cref{tab:abla_dm} (a), (b), \cref{tab:alpha_effect} and \cref{fig:abla_diff}.
From \cref{tab:abla_dm} (a), we first validate the advantage of performing $y_0$-prediction over predicting noise.
From \cref{tab:abla_dm} (b), we confirm that our proposed scheduler is the best choice for scheduling with the best overall results, as it ensures the $y_0$-prediction task remains non-trivial throughout the entire denoising chain.
\Cref{tab:alpha_effect} further supports our hypothesis regarding the effects of $\bar{\alpha}_0$, demonstrating that setting $\bar{\alpha}_0 = 0.5$ enhances both prediction accuracy and fidelity.
From \cref{fig:abla_diff}, we validate that $T=10$ is a reasonable choice for number of diffusion steps with its best overall performance.
Crucially, as an single-stage learning framework, \name~achieves state-of-the-art performance using only a minimal number of diffusion steps, in contrast to previous methods \cite{chen2023humanmac, DBLP:conf/aaai/WeiSLLLSH23}. 
As indicated in the bottom rightmost sub-figure of \cref{fig:abla_diff} with its low inference time, \name~is highly efficient and easy to optimize.

\begin{table}[t]\scriptsize
\centering
\caption{\textbf{Left (a)}: Ablation on loss configurations. 
$\gamma = 0$ means that the model does not try to reconstruct the motion history $x$. 
$\lambda^j = 1$ means that all joints are weighted equally. \textbf{Right (b)}: Ablation on implicit diversity relaxation.}
\resizebox{\linewidth}{!}{
\begin{tabular}{c | c c c c c}
\hline \multicolumn{1}{c}{Loss} & APD $\uparrow$ & ADE $\downarrow$ & MMADE $\downarrow$ & CMD $\downarrow$ & FID $\downarrow$ \\
\hline 
 $\gamma = 0$ & 7.661 & 0.351 & 0.496 & \textbf{2.742} & 0.123 \\
 $\lambda^j = 1$ & 7.609 & 0.352 & 0.494 & 2.970 & 0.115 \\
 $\ell_2$ & \textbf{9.054} & 0.378 & 0.509 & 8.215 & 0.204 \\
 \rowcolor{gray}
 ours & 7.632 & \textbf{0.350} & \textbf{0.494} & 3.202 & \textbf{0.102} \\
\hline
\end{tabular}
\begin{tabular}{c | c c c c c}
\hline \multicolumn{1}{c}{$k$} & APD $\uparrow$ & ADE $\downarrow$ & MMADE $\downarrow$ & CMD $\downarrow$ & FID $\downarrow$ \\
\hline 
1 & 4.061 & \textbf{0.346} & 0.503 & 6.094 & 0.163 \\
\rowcolor{gray}
2 (ours) & 7.632 & 0.350 & \textbf{0.494} & \textbf{3.202} & \textbf{0.102} \\
3 & \textbf{9.233} & 0.372 & 0.506 & 5.036 & 0.191 \\
\hline
\end{tabular}
}
\label{tab:combined_ablation}
\end{table}

\subsubsection{Loss Configuration.}
The designed loss $\gL_{\textrm{final}}$ is essential to \name~'s performance.
As such, we investigate (1) the components of \cref{eq::total_loss} and (2) the implicit diversity relaxation technique.
First, from \cref{tab:combined_ablation} (a), we find that both reconstructing $x$ and using structural weights contribute to \name's performance. Moreover, while the $\ell_1$ loss may offer sub-optimal diversity, it excels in terms of accuracy and fidelity when compared with the $\ell_2$ loss.
Second, by varying the relaxation hyperparameter $k$ as described in \cref{eq::loss_relax} and analyzing the results in \cref{tab:combined_ablation} (b), we observe that setting $k$ to 2 emerges as the optimal choice for most metrics, which helps \name~maintaining a good balance between sample diversity, accuracy and fidelity.
\section{Conclusion}
\label{sec::con}

In this work we present~\name, a novel end-to-end DM-based framework for stochastic HMP.
Benefiting from the GCN-DCT design used in deterministic works, \name~addresses issues of previous methods as it produces realistic, behaviorally consistent, and properly diverse human motions through single-stage learning.
The motion predictor of \name~features a Transformer-based reconstruction module and a GCN-based refinement module, collaboratively learning future motion from its corrupted form and the provided motion history. 
By predicting motion directly using this dual design instead of noise and a simple yet effective variance scheduler, \name~establishes a new paradigm in stochastic HMP. 
The results obtained from extensive experiments and analyses confirm that \name~achieves significant performance gains over state-of-the-art baselines on benchmark datasets, demonstrating the efficacy of our method.
\newpage
\section*{Acknowledgements}

This work is supported in part by Navy N00014-19-1-2373, the joint NSF-USDA CPS Frontier project CNS \#1954556, USDA-NIFA \#2021-67021-34418, and Agriculture and Food Research Initiative (AFRI) grant no. 2020-67021-32799/project accession no.1024178 from the USDA National Institute of Food and Agriculture: NSF/USDA National AI Institute: AIFARMS. 
Work is also supported in part by NSF MRI grant \#1725729 \cite{10.1145/3311790.3396649}.

%
%
\bibliographystyle{splncs04}
\bibliography{egbib}
\newpage
\appendix
\renewcommand{\thefigure}{A.\arabic{figure}}
\renewcommand{\thetable}{A.\arabic{table}}
\setcounter{figure}{0}
\setcounter{table}{0}
\renewcommand{\theequation}{A.\arabic{equation}}
\setcounter{equation}{0}

\section{Weights $\lambda^i$ Derivation Details}
\label{subsec::lambda}
The structure-aware reconstruction loss is defined as follows:
\begin{equation}
\label{eq::total_loss_app}
\gL_{\textrm{rec}}=\frac{1}{J}\sum_{j=1}^J(\gamma \cdot \|(x^j-\hat{x}^j) \cdot \lambda^j\|_1 + \|(y_0^j-\hat{y}_0^j) \cdot \lambda^j\|_1),
\end{equation}
where $x^j$ and $\hat{x}^j$ represent the groundtruth and predicted positions of the $j^{th}$ joint in the motion history, respectively. 
Similarly, $y_0^j$ and $\hat{y}_0^j$ denote corresponding values in the target future motion. 
The superscript $j$ indicates the joint index.
In this formulation, the notations $x, y \in \sR^{J\times 3}$ represent a single pose containing $J$ joints, and the loss is averaged across the temporal dimension of the pose sequences.

Our weight assignment method for $\lambda^j$, which draws inspiration from the approach described in \cite{DBLP:journals/corr/abs-2305-04443}, is based on the kinematic structure of the human body. 
Kinematic chains, defined as a series of linked joints from the base to the end joint, are vital in human motion modeling due to their depiction of joint connectivity and movement dynamics. 

Formally, each pose $x$ or $y$ is described by $L$ such chains. 
Let $c_l$ be the $l^{th}$ kinematic chain, $b_l^{i}$ the bone length of the $i^{th}$ bone on $c_l$, and $l(c_l)$ the total number of bones in $c_l$. 
For a joint $x^j$ or $y^j$ that is the $j'^{th}$ joint on chain $c_l$, the weight $\lambda^j$ is computed as follows:
\begin{align}
\label{eq::lambda_t_app}
&\lambda^j \propto \frac{j'}{l(c_l)}\ln\left(\sum_{i'=1}^{j'}b_l^{i'}\right), \\
&\sum_{j=1}^{J}\lambda^j = 1.
\end{align}
This weighting scheme assigns higher weights to dynamically active joints, typically external ones, acknowledging their significant contribution to the quality of the predicted human motion.

\section{CMD and APDE Derivation Details}

CMD (Cumulative Motion Distribution) and APDE (Average Pairwise Distance Error) are two new metrics proposed in \cite{barquero2023belfusion} for evaluating stochastic HMP models. 
We outline their derivation details below, illustrating their significance in capturing key aspects of motion fidelity and diversity.

CMD measures the difference between the areas under the cumulative true motion and predicted motion distributions.
Let $\bar{M}$ denote the $\ell_2$ distance between joint coordinates in two consecutive frames (displacement) across the entire test partition of the dataset. 
For the $f^{th}$ frame in all predicted motions, we compute
the average displacement $M_f$.
The overall CMD is then computed as:

\begin{align}
\textrm{CMD} &= \sum_{i=1}^{F-1} \sum_{f=1}^i\left\|M_f-\bar{M}\right\|_1 \\
&=\sum_{f=1}^{F-1}(F-f)\left\|M_f-\bar{M}\right\|_1,
\end{align}
where $F$ represents the total number of predicted frames.
Frame-wise CMD, illustrated in Fig. 3 of the main paper, is computed for each frame $i$ as:
\begin{equation}
\textrm{CMD}(i) = \sum_{f=1}^{i}(i-f+1)\left\|M_f-\bar{M}\right\|_1, i \in [1, F-1],
\end{equation}
where $i$ is the frame index.
This frame-wise analysis provides a deeper insight into the motion fidelity up to each specific point in time throughout the predicted trajectory.

APDE quantifies the error between the APD (Average Pairwise Distance) of the multimodal groundtruth and the predictions.
For each set of predicted samples $\{\hat{y}\}$, APDE is calculated as:
\begin{equation}
\textrm{APDE} = \lvert \textrm{APD}_{y}-\textrm{APD}(\{\hat{y}\})\rvert,
\end{equation}
where $\textrm{APD}_{y}$ represents the APD of the multimodal groundtruth for $y$ obtained by grouping similar past motions.
This metric effectively captures the deviation of the diversity of the predicted motion from the expected diversity in the groundtruth, measuring to what extent the diversity is properly modeled.


\section{\name~Implementation Details}

\subsection{General Settings}
We train \name~as a 10-step DM with standard DDPM \cite{DBLP:conf/nips/HoJA20} sampling.
Adam~\cite{DBLP:journals/corr/KingmaB14} is used for all experiments with 0.0001 as the initial learning rate.
Training batch size are 64 and 32 for Human3.6M and AMASS respectively.
We use PyTorch \cite{paszke2019pytorch} to implement \name, and experiments are conducted with NVIDIA V100 and A100 GPUs.

\subsection{Motion Reconstruction Module $F(\cdot)$}
The motion reconstruction module $F(y_t, t)$ aims to generate an ``initial reconstruction'' $\Tilde{y}_0$ from the target noisy motion $y_t$, and this reconstruction is independent of the motion history $x$. 
$F(\cdot)$ is composed of $8$ transformer encoder layers, each with $4$ attention heads, a latent dimension of $512$, and a dropout ratio of $0.1$.
The feedforward layer in each transformer layer has a dimension of $1024$.
Both the time step $t$ embedding and the positional encoding are sinusoidal, which are used to obtain temporal information across the denoising chain and the motion trajectory respectively.
We use a $2$-layer MLP to project the time step $t$ embedding to the transformer latent dimension.
GELU activation is used throughout the motion reconstruction module.

\subsection{Motion Refinement Module $R(\cdot)$} 
The motion refinement module $R(x, \Tilde{y}_0)$ aims to reconstruct the entire motion trajectory guided by the motion history $x$.
$R(x, \Tilde{y}_0)$ consists of $3$ blocks, each of which contains $2$ GCN-based
residual layers. 
Within these blocks, motion sequences are initially converted into DCT (Discrete Cosine Transform) coefficients, processed in the frequency domain, and then projected back into the pose space.
Each residual layer contains $2$ GCN layers followed by batch normalization layers.
The latent dimension and the dropout ratio of the refinement module are $256$ and $0.5$.
Tanh activation is used throughout the motion refinement module.

\begin{wrapfigure}[16]{R}{0.5\textwidth}
\centering
\includegraphics[width=0.5\textwidth]{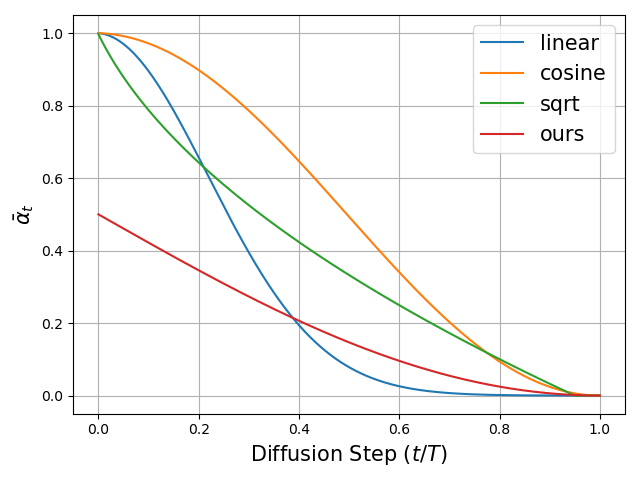}
\caption{Values of $\bar{\alpha}_t$ throughout diffusion in the linear scheduler, cosine scheduler, sqrt scheduler and ours. Recall that $\bar{\alpha}_{t} = \prod_{s=1}^{t} \alpha_s$.}
\label{fig:sched_app}
\end{wrapfigure}

\subsection{Variance Scheduler $\{1 - \alpha_t\}_{t=0}^T$}
Our variance scheduler is derived by modifying the original cosine scheduler as follows:
\begin{equation}
\label{eq::scheduler_app}
    \bar{\alpha}_t = \cos \left(\frac{t / T+ 1}{2} \cdot \frac{\pi}{2}\right)^2.
\end{equation}
As outlined in the main paper, we establish an initial value of $\bar{\alpha}_0 = 0.5$ by setting the offset to $1$, deviating from the traditional $\bar{\alpha}_0 \approx 1$ approach.
A visual comparison of our proposed variance scheduler with the standard linear, cosine, and sqrt schedulers is presented in \cref{fig:sched_app}, illustrating the distinctions of our approach.

\subsection{Loss Configuration} 
As detailed in \cref{subsec::lambda}, the weights $\lambda^j$ are pre-computed based on the details of the kinematic structure of the human body. 
They can be derived from the input data before training \name, thus only imposing minimal computational overhead. 
We set $\gamma = 1/10$ in \cref{eq::total_loss_app} to balance the importances of the reconstruction of motion history and the prediction of future motion.
As mentioned in the main paper, the implicit diversity relaxation parameter $k$ is set to $2$.

\subsection{Datasets} 

In our dataset setup, we adhere to the protocol established in \cite{barquero2023belfusion} to ensure a fair comparison with prior works.
For the Human3.6M dataset, we utilize $37,133$ motion samples for training and $5,168$ for evaluation. 
In the case of the AMASS dataset, $120,758$ motion samples are used for training and $12,727$ for evaluation. 
The data augmentation technique from \cite{barquero2023belfusion} is also adopted, where all pose sequences are randomly rotated from $0$ to $360$ degrees around the $Z$-axis during training. 
Additionally, the $\ell_2$ distance thresholds for generating the multimodal groundtruth for each data sample are set to $0.5$ for Human3.6M and $0.4$ for AMASS.

\subsection{General Learning Setting}
In line with previous studies, we set the number of history frames $H$ and future frames $F$ as $H=25$, $F=100$ for the Human3.6M dataset, and $H=30$, $F=120$ for the AMASS dataset. Consequently, the number of DCT coefficients used in the motion refinement module $R(\cdot)$ is adjusted to $125$ for Human3.6M and $150$ for AMASS, corresponding to the total number of frames in the motion trajectories of each dataset.
Due to GPU memory constraints, the training batch sizes are configured as $64$ for Human3.6M and $32$ for AMASS. \name~is trained over $500$ epochs for both datasets, with the learning rate starting at $0.0001$ and beginning to decay after the $200^{th}$ epoch. For reproducibility and consistency across experiments, we use a random seed of $0$.

\section{Effeciency of \name: Space and Time Comparison}

\begin{table}[t]\footnotesize	
\centering
\caption{Inference time and parameter quantity comparison between \name~and state-of-the-art methods on Human3.6M dataset.}
\begin{tabular}{c|c|cc}
\hline 
\multicolumn{1}{c}{Type} & \multicolumn{1}{c}{Method} & Inference Time & Parameter Quantity \\
\hline 
\multirow{3}{*}{VAE-based} & DLow \cite{DBLP:conf/eccv/YuanK20} & 0.314s & 8.017M \\
& GSPS \cite{DBLP:conf/iccv/MaoLS21} & 0.012s & 1.298M \\
& DivSamp \cite{DBLP:conf/mm/DangNLZL22} & 0.025s & 23.102M \\
\hline 
\multirow{3}{*}{DM-based} &
BeLFusion \cite{barquero2023belfusion} & 28.802s & 13.193M \\
& HumanMAC \cite{chen2023humanmac} & 1.255s & 28.402M \\
 & \cellcolor{gray}ours & \cellcolor{gray}0.179s & \cellcolor{gray}18.790M \\
\hline
\end{tabular}
\label{tab:space_time}
\end{table}

In \cref{tab:space_time}, we compare the inference time and parameter quantity of \name~\\with other state-of-the-art methods. 
This comparison is essential for understanding the practical efficiency and scalability of our model. 
The inference time, crucial for real-time applications, is measured by the time taken to generate $50$ motion samples from a single motion history.
As shown in \cref{tab:space_time}, despite its relatively high parameter quantity, \name~achieves a generation speed comparable to efficient VAE-based methods.
Notably, \name~exhibits a significantly higher efficiency, surpassing other DM-based methods, BeLFusion \cite{barquero2023belfusion} and HumanMAC \cite{chen2023humanmac} \footnote{The inference time of HumanMAC is obtained through its $100$-step DDIM \cite{song2020denoising} (original $1000$-step diffusion chain) sampling.}, by orders of magnitude. 
This enhanced efficiency is primarily due to \name's short denoising chain and its RNN-free motion generator architecture, which collectively contribute to its faster processing capabilities.

\section{Additional Qualitative Results}

This section introduces supplementary images and videos, available in respective directories, to showcase \name's ability to produce future motion sequences that are not only realistic but also consistent with the given motion history.
These additional materials emphasize \name's capacity to strike a balance between sample fidelity and diversity.

\subsection{Images} 
The images are located in the \texttt{h36m\_imgs} and \texttt{amass\_imgs} sub-folders.
Each image, named as \texttt{[class\_name]\_[sample\_id]\_[dataset\_name].pdf}, showcases prediction comparisons for a single, randomly sampled motion history from a specific class (sub-dataset) of either the Human3.6M or AMASS dataset. 
The images display the motion history ($0.5$ seconds) in the first $3$ poses (\textit{green-purple}) and the predicted future motion ($2$ seconds) in the subsequent $4$ poses (\textit{blue-orange}), with $10$ predicted samples overlaid on the groundtruth.
From top to bottom, the images include results from DLow \cite{DBLP:conf/eccv/YuanK20}, GSPS \cite{DBLP:conf/iccv/MaoLS21}, DivSamp \cite{DBLP:conf/mm/DangNLZL22}, BeLFusion \cite{barquero2023belfusion} and \name, mirroring the layout in Fig. 4 of the main paper. 
These visualizations highlight \name's ability to generate properly diverse motions with minimal anomalies which are physically implausible.

\subsection{Videos} 
The videos, available in the \texttt{h36m\_mp4s} and \texttt{amass\_mp4s} sub-directories, are named similarly to the images and have one-to-one correspondence with them. 
Focusing solely on \name, each video shows the motion history context in the first column, the groundtruth motion sequence in the second, and $5$ \name~predictions in the remaining columns.
These videos serve as a further proof of \name's ability to produce visually consistent and natural human motion.

\section{Additional Ablation Studies}

This section provides additional ablation results and analyses that were not included in the main paper due to space constraints. 
Conducted using the Human3.6M dataset, these studies further clarify the impact of each design choice on \name's overall performance.

\subsection{General Framework Components}

\begin{table*}[t]\scriptsize	
\centering
\caption{Full ablation on \name's general architecture. In the \texttt{\textit{Sched.}} column, \cmark~denotes use of our proposed scheduler.} 
\resizebox{\linewidth}{!}{
\begin{tabular}{c c c | c c c c c c c c}
\hline $F(y_t, t)$ & $R(x, \Tilde{y}_0)$ & \multicolumn{1}{c}{\texttt{\textit{Sched.}}} & APD $\uparrow$ & APDE $\downarrow$ & ADE $\downarrow$ & FDE $\downarrow$ & MMADE $\downarrow$ & MMFDE $\downarrow$ & CMD $\downarrow$ & FID $\downarrow$\\
\hline 
\xmark & \xmark & \xmark & 12.880 & 4.854 & 0.959 & 1.000 & 0.987 & 1.004 & 966.716 & 1.047 \\
\xmark & \xmark & \cmark & \textbf{24.452} & 16.367 & 1.724 & 1.467 & 1.737 & 1.468 & 2408.312 & 2.760 \\
\xmark & \cmark & \cmark & 7.588 & 1.679 & 0.498 & 0.607 & 0.583 & 0.628 & 259.037 & 0.680	\\
\xmark & \cmark & \xmark & 6.858 & 1.835 & 0.539 & 0.678 & 0.625 & 0.694 & 197.105 & 0.474 \\
\cmark & \xmark & \xmark & 3.727 & 4.441 & 0.502 & 0.669 & 0.632 & 0.731 & \textbf{3.176 }& 0.167	 \\
\cmark & \xmark & \cmark & 3.835 & 4.338 & 0.494 & 0.653 & 0.628 & 0.721 & 3.382 & 0.193 \\
\cmark & \cmark & \xmark & 7.602 & \textbf{1.446} & 0.382 & 0.489 & 0.521 & 0.537 & 3.323 & 0.282 \\
\rowcolor{gray}
\cmark & \cmark & \cmark & 7.632 & 1.609 & \textbf{0.350} & \textbf{0.458} & \textbf{0.494} & \textbf{0.506} & 3.202 & \textbf{0.102} \\
\hline
\end{tabular}}

\label{tab:abla_gene_app}
\end{table*}

We present the full evaluation of the individual contributions of \name's components in \cref{tab:abla_gene_app}.
The comprehensive experimental results yield the following additional insights:
(1) The motion reconstruction module $F(\cdot)$ provides crucial guidance for the model in producing accurate and consistent predictions that closely align with the motion history. 
From the table, the absence of $F(\cdot)$ leads to a significant increase in the CMD score, indicating large displacements among the predicted poses.
(2) The motion refinement module $ R(\cdot) $ plays a key role in ensuring an appropriate level of diversity in the predicted motion samples. 
This is reflected in lower APDE and decreased MMADE and MMFDE when compared to variants excluding $R(\cdot)$.
(3) The proposed variance scheduler is helpful in improving \name's overall performance, underscoring its effectiveness.
In summary, the results highlight the importance of the collaborative function of \name's components in achieving its performance.

\subsection{Number of Transformer Encoder Layers Used in $F(\cdot)$}

\begin{table}[t]\footnotesize	
\centering
\caption{Ablation on number of transformer encoder layers used in $F(\cdot)$.} 

\begin{tabular}{c|ccccc}
\hline 
\multicolumn{1}{c}{\# Layers} & APD $\uparrow$ & ADE $\downarrow$ & MMADE $\downarrow$ & CMD $\downarrow$ & FID $\downarrow$ \\
\hline 
 1 & \textbf{8.682} & 0.371 & 0.515 & 3.450 & 0.177 \\
 2 & 8.633 & 0.364 & 0.507 & 3.386 & 0.180 \\
 4 & 7.888 & 0.358 & 0.499 & \textbf{2.892} & 0.133 \\
 \rowcolor{gray}
 8 (ours) & 7.632 & \textbf{0.350} & \textbf{0.494} & 3.202 & \textbf{0.102} \\
 10 & 7.768 & 0.355 & 0.495 & 3.447 & 0.124 \\
\hline
\end{tabular}
\label{tab:abla_layer}
\end{table}

In \cref{tab:abla_layer}, we present the results of experiments conducted with different numbers of transformer encoder layers in the motion reconstruction module $F(\cdot)$. 
Based on these results, we choose to use $8$ layers for $F(\cdot)$, as this configuration offers the best overall performance.

\subsection{GCN Configuration of $R(\cdot)$}

\begin{table}[t]\footnotesize	
\centering
\caption{Ablation on number of GCN blocks used in $R(\cdot)$.} 
\begin{tabular}{c|ccccc}
\hline 
\multicolumn{1}{c}{\# Blocks} & APD $\uparrow$ & ADE $\downarrow$ & MMADE $\downarrow$ & CMD $\downarrow$ & FID $\downarrow$ \\
\hline 
 1 & 7.718 & 0.367 & 0.496 & \textbf{2.758} & 0.181	 \\
 2 & 7.668 & 0.356 & 0.496 & 2.990 & 0.120	 \\
\rowcolor{gray}
 3 (ours) & 7.632 & 0.350 & \textbf{0.494} & 3.202 & \textbf{0.102} \\
 6 & \textbf{7.731} & \textbf{0.346} & 0.497 & 3.647 & 0.115	\\
\hline
\end{tabular}
\label{tab:abla_gcn}
\end{table}

In \cref{tab:abla_gcn}, we present the results from experiments that explored using different numbers of GCN blocks in the motion refinement module $ R(\cdot) $. 
The results demonstrate an increase in the accuracy of predicted samples when more GCN blocks are used, as evidenced by a consistent decrease in ADE. 
This trend confirms the effectiveness of employing explicit spatial-temporal modeling through GCN for motion data. 
Based on these results, we opt to implement $3$ GCN blocks in $R(\cdot)$ to achieve the best overall performance.

\subsection{Number of DCT Coefficients of $R(\cdot)$}

\begin{table}[t]\footnotesize	
\centering
\caption{Ablation on number of DCT coefficients used in $R(\cdot)$.} 
\begin{tabular}{c|ccccc}
\hline 
\multicolumn{1}{c}{\# DCT coef.} & APD $\uparrow$ & ADE $\downarrow$ & MMADE $\downarrow$ & CMD $\downarrow$ & FID $\downarrow$ \\
\hline 
 10 & \textbf{8.235} & 0.386 & 0.489 & 3.783 & 0.124	 \\
 20 & 8.072 & 0.374 & 0.495 & 2.992 & 0.152	 \\
 50 & 7.999 & 0.360 & 0.494 & \textbf{2.751} & 0.123	 \\
 100 & 7.799 & 0.355 & 0.495 & 2.786 & 0.105	\\
 \rowcolor{gray}
 125 (ours) & 7.632 & \textbf{0.350} & \textbf{0.494} & 3.202 & \textbf{0.102} \\
\hline
\end{tabular}
\label{tab:abla_dct}
\end{table}

Previous works such as \cite{DBLP:conf/iccv/MaoLSL19, chen2023humanmac} demonstrate that using a subset of DCT coefficients can lead to better motion prediction performance while achieving better computational cost.
To this end, we study the effect of number of DCT coefficients used in $R(\cdot)$.
From \cref{tab:abla_dct}, we observe that \name~requires all DCT coefficients to achieve the best performance.

\end{document}